\documentclass[sn-mathphys]{sn-jnl}

\jyear{2022}%

\theoremstyle{thmstyleone}%
\theoremstyle{thmstyletwo}%

\theoremstyle{thmstylethree}%
\usepackage{changebar}
\begin{document}


\title{Human Gait Recognition using Deep Learning: A Comprehensive Review}

\author[1]{\fnm{Muhammad Imran} \sur{Sharif}}\email{imransharif@ksu.edu}

\author[2]{\fnm{Mehwish} \sur{Mehmood}}\email{mmehmood01@qub.ac.uk}

\author[3]{\fnm{Muhammad Irfan} \sur{Sharif}}\email{irfan.sharif@ue.edu.pk}

\author[4]{\fnm{Md Palash} \sur{Uddin}}\email{m.uddin@deakin.edu.au}

\affil[1]{\orgdiv{Department of Computer Science, Kansas State University, Manhattan, KS 66506, USA}}

\affil[2]{\orgdiv{School of Electronics, Electrical Engineering and Computer Science, Queen's University Belfast, United Kingdom}}

\affil[3]{\orgdiv{Department of Computer Science, University of Education, Jauharabad Campus, Khushāb 41200, Pakistan}}

\affil[4]{\orgdiv{School of Information Technology, Deakin University, Geelong, VIC 3220, Australia}}


\abstract{Gait recognition (GR) is a growing biometric modality used for person identification from a distance through visual cameras. GR provides a secure and reliable alternative to fingerprint and face recognition, as it is harder to distinguish between false and authentic signals. Furthermore, its resistance to spoofing makes GR suitable for all types of environments. With the rise of deep learning, steadily improving strides have been made in GR technology with promising results in various contexts. As video surveillance becomes more prevalent, new obstacles arise, such as ensuring uniform performance evaluation across different protocols, reliable recognition despite shifting lighting conditions, fluctuations in gait patterns, and protecting privacy.This survey aims to give an overview of GR and analyze the environmental elements and complications that could affect it in comparison to other biometric recognition systems. The primary goal is to examine the existing deep learning (DL) techniques employed for human GR that may generate new research opportunities.}

\maketitle

\section{Introduction}\label{sec1}

Gait Recognition (GR) is a method of identifying individuals by their walking patterns \cite{alom2019state}. There are different types of human recognition, such as face recognition, iris recognition, voice recognition, hand geometry, signature, vein, and gait recognition \cite{shah2014face,murtaza2014face,murtaza2013analysis,sharif2012single,aisha2014face,sharif2019overview}. DL algorithms have been widely used in various biometrics tasks such as human face and iris recognition \cite{de2014firme}. However, very few studies focus on GR using DL.\\\\
GR is a challenging issue due to the subtle differences in people's walk \cite{wang2003silhouette}. Computers find it challenging to recognize human gaits because subtle movements are hard to detect and variations between people, body types, and conditions \cite{zhang2007human}. DL has significantly improved many image recognition tasks, like object detection or image segmentation \cite{ghaffarian2021effect}. It is also showing promising results in GR. GR is used for identification and authentication applications such as user authentication, access control \cite{gafurov2010improved}, etc. It is a biometric measure that captures and analyzes gait patterns from video or images to authenticate a person \cite{kumar2021gait}. A GR system can be used for two different purposes: which are discussed below.

\subsection{Identification based on Gait Recognition}\label{subsec1}
In this use case, a system has to verify if a person is the same person who has taken earlier from the database sample \cite{wang2022human}. The sample may be any video or images of the person’s gait and the features of a gait are required to match with an image or a video \cite{chao2021gaitset}. To make it more accurate, the gait samples can be taken from various conditions like walking speed, walking on different surfaces, taking samples while carrying baggage, under conditions like indoors or outdoors, etc \cite{mu2021resgait}. The robustness of the model is dependent on the number of samples and variations of the samples collected to train the system. A more robust model is more likely to identify the same person consistently even when walking at different speeds, carrying different types of luggage \cite{filipi2022gait}, etc.\\

\subsection{Verification based on Gait Recognition}
Unlike identification, where the system has to identify a person, in verification \cite{batool2020offline}, the system has to match a sample that is taken at one time with another sample that is taken at other times. This is more challenging because the system has to account for all factors that impact an individual's gait, such as mood, age, health, fitness level, etc \cite{topham2022human}.\\
Fig \ref{fig.1} below depicted the human GR block diagram. 

 \begin{figure}[H]%
		\centering
		\includegraphics[width=1.1\textwidth]{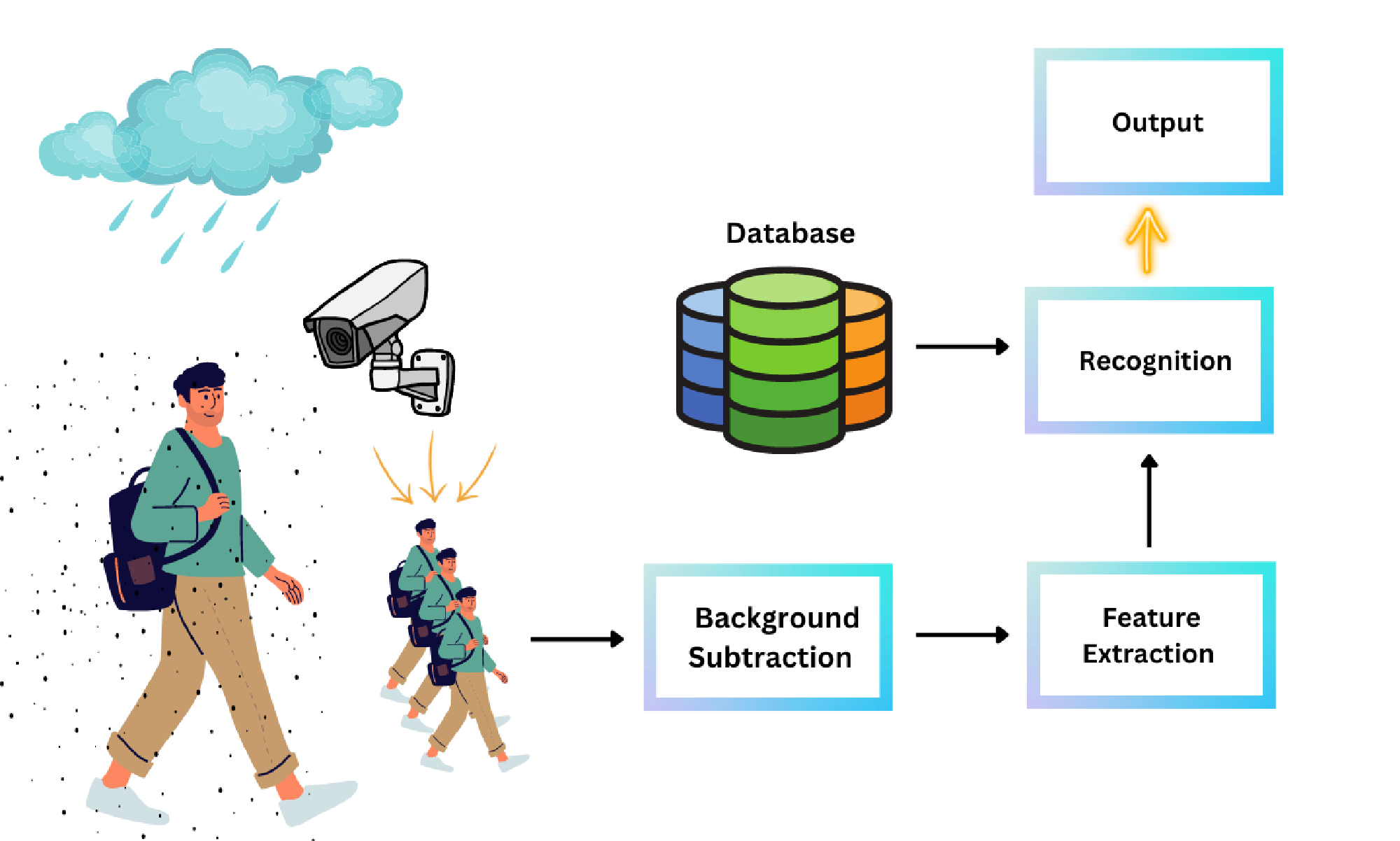}
		\caption{Human GR block diagram \cite{hayder2011person}. The output is obtained by using background subtraction to pass through the video or images of a moving person, extracting features from the provided data, and then comparing the resulting images to those in the database for recognition.}\label{fig.1}
	\end{figure}
GR systems are beneficial when other biometric systems, such as fingerprint recognition or facial recognition, do not work well because of their ineffectiveness in challenging environments \cite{shen2022comprehensive}.\\

\section{Overview of Gait Recognition}

GR has appeared in science fiction for decades, from the automated doors in Star Trek to the futuristic police identification systems in Minority Report \cite{dinello2021six}. Despite that, the technology to perform GR was not available or reliable until the last decade \cite{alobaidi2022real}. Visual pattern recognition is a challenging job in computer science \cite{agrawal2021specific}. It is also subject to variations and changes depending on the person, the environment, and the time of day \cite{zhu2021gait}. It has been performed with a variety of different techniques \cite{topham2022gait}. Some of the earliest work was based on computer vision and analyzing the movement of limbs or body shapes, while more recent work has used acoustic techniques \cite{jiang2022utilizing}.\\

\subsection{Distinctive Properties of Gait Recognition}
GR has distinctive properties that make it challenging as compared to other biometrics \cite{dou2022gaitmpl}. Due to the variation in gait, it is an unusual biometric and is often referred to as a “fool-proof” biometric \cite{sindhu2022personnel}. Additionally, GR cannot be modified aside from paralysis or amputation
\cite{isaac2021robust}. Furthermore, since individuals walk daily, the data used for ML algorithms can be collected easily and GR is classified as a passive biometric requiring no active user effort \cite{lopes2022gait}. \cite{parashar2022intra}. These factors make GR an especially valuable biometric.\\

\subsection{Importance of Gait Recognition}
Gait Recognition (GR) is a valuable tool in various situations and uses, yet it can be difficult for computers to perform this task effectively \cite{song2022casia}. Furthermore, humans are hard enough to recognize from videos or pictures, making GR even more challenging \cite{han2022unified}. It consists of using data obtained through visual sensors such as cameras and then collecting and processing the information to identify the individual's gait \cite{baghezza2021profile}.This is made complicated due to the two-dimensional nature of visual data. That means the computer only has two visuals to work with. Additionally, GR tends to vary slightly depending on who is performing it and the conditions they are walking in, so computers must be able to recognize minor alterations while keeping up a high degree of accuracy \cite{wei2022coupled}. \\

\subsection{Human Gait Representation}
\begin{figure}[H]%
		\centering
		\includegraphics[width=1.1\textwidth]{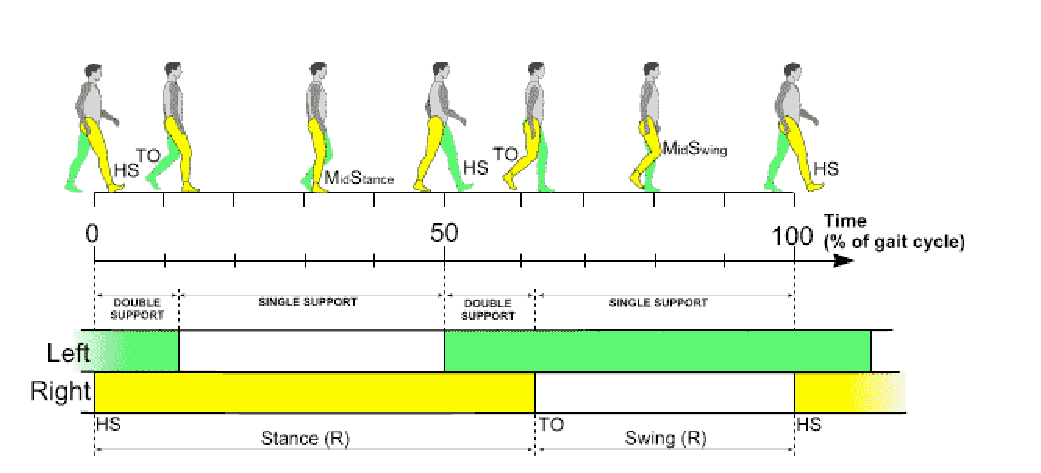}
		\caption{Human gait cycle between stance to swing phase \cite{mobasseri2009time}. The green color shows the movement of the left leg and the yellow color shows the moment of the right leg. We can also see single support in which the person is getting support from a single leg while moving and double support in which both legs of the person are on the ground.}\label{fig.2}
	\end{figure}

The human gait cycle is a series of phases in which the body moves from one position to another \cite{bijalwan2021fusion}. The stance phase starts with the flat foot on the ground and ends with the body in a fully upright position \cite{song2021compliance}. The swing phase follows when the body is in motion. The additional phases include toe-off and heel-off, mid-swing, final swing, and pause \cite{semwal2021speed}. Most sports and activities aim to move from one point to another as quickly and efficiently as possible. Our human gait cycle comprises two significant phases: stance and swing which are discussed below. \\\\
\begin{enumerate}
    \item \textbf{Stance Phase}: The stance phase lasts from just before foot contact to when the ball of your foot hits the ground again after toe-off. This time allows you to prepare your muscles and joints for movement by stretching and lengthening them.\\\\
 
  \item \textbf{Swing Phase}: The swing phase starts immediately after foot contact and ends when your ankle comes back under control, or you lift off the ground by pushing off with your toes. During this time, you are using the small muscles in your legs to push against the ground with each step you take. \\\\
  The cycle of these two phases is depicted in Fig \ref{fig.2}.
 \end{enumerate}

\section{Environmental Factors that Affect Gait Recognition}

An individual's gait is affected by many factors, including lighting, hearing, stress levels, exhaustion and diseases like Parkinson’s \cite{rahman2008factors}. If you plan to implement GR in your business or organization, you must understand how various environmental factors may impact its effectiveness \cite{silva2020basics}. Several environmental factors can affect the precision of GR \cite{sprager2015inertial}. These include: \\\\
\begin{enumerate}
    \item Lighting Conditions - GR is 25\% less accurate in low-light conditions than in bright indoor light. \\
 \item	Scarring - Traditionally, GR is used to identify and distinguish suspects in criminal proceedings. If an individual has a distinctive scar not located on the side of their face, it can be employed as a form of identification. \\
 \item	Clothing - Certain types of apparel, such as large bags or backpacks, can negatively influence the accuracy of GR. \\
 \item Environmental Conditions - GR is more precise when used indoors and less precise when utilized outdoors.\\\\
 
\subsection{Security Strength of Gait Bio-metric} 
GR can be valid for security and medical purposes, like recognizing people who enter a facility or monitoring a patient's progress\cite{muro2014gait}. Additionally, it can distinguish persons in crowded areas, which may be advantageous for law enforcement, intelligence gathering, and other safety-related domains\cite{alobaidi2022real} \cite{shahid2022data}. Furthermore, GR is usually more precise in indoor settings with sufficient lighting than in outdoor and dimly lit conditions\cite{dai2021development}.\\\\
A GR algorithm must be trained on many subjects' walking styles before accurately identifying people \cite{alsaggaf2021smart}. Using more subjects than fewer subjects to train the algorithm is better. This means that the algorithm is trained on various walking patterns and can identify various people. GR can be used as a bio-metric modality as it is not easily spoofed \cite{sepas2022deep}. There have been instances where people try to fake a person's gait, but these have been unsuccessful. This is because learning how to walk like another person is challenging \cite{vildjiounaite2006unobtrusive}.\\

\end{enumerate}

\section{Gait Recognition Methods}
\begin{figure}[H]%
		\centering
		\includegraphics[width=1.1\textwidth]{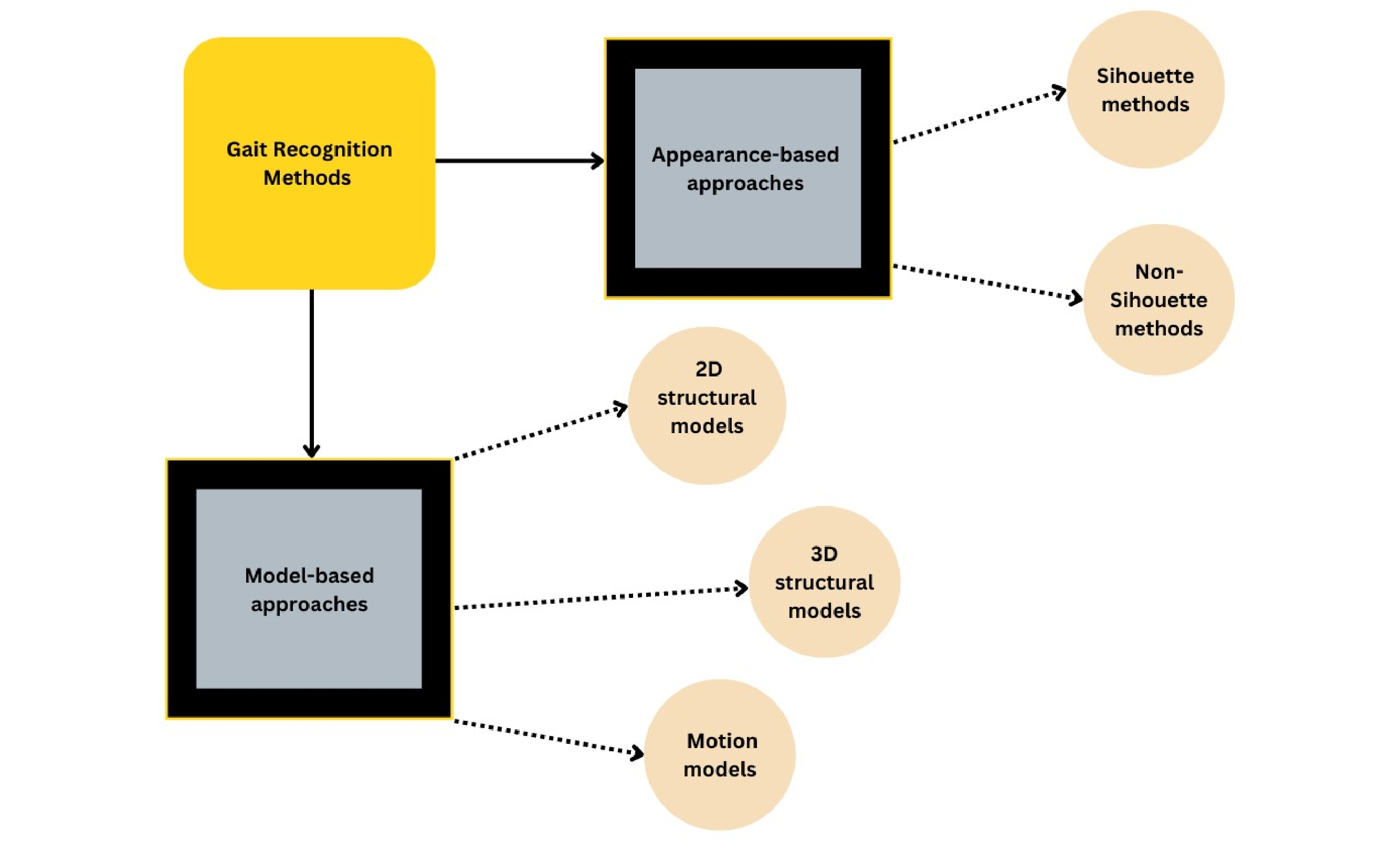}
		\caption{GR methods are divided into appearance and model-based approaches \cite{bouchrika2018survey}. The first one is further divided into two methods; sihouette methods and non-sihouette methods. The later is divided into three models; 2D structural models, 3D structural models and motion models.}\label{fig.3}
	\end{figure}

Human GR is a field of research at an early-to-mid stage \cite{kirtley2006clinical}. The computational complexity of most methods remains high, however. For example, many approaches require multiple cameras and viewpoint normalization. Most methods still have some room for improvement. Regardless of their limitations \cite{teepe2021gaitgraph}, these technologies have the potential to revolutionize gait analysis. To perform the GR task, various types of approaches including appearance-based and model-based approaches are depicted in Fig \ref{fig.3}. 

\begin{enumerate}
    \item Appearance-based approaches for object recognition seek to find objects by their visual appearance \cite{leibe2003analyzing}. This contrasts with other methods that attempt to find objects by their geometrical, physical, or functional properties. These methods are generally computationally intensive and are thus often used in combination with other methods to create a more efficient object recognition system. \\\\

   \item Model-based approaches have become increasingly popular in a variety of different fields \cite{zhong2021sequential}. These approaches are particularly effective in the domain of ML. There are several advantages to using model-based approaches. These advantages include the ability to improve generalization, the ability to capture relationships between variables, and the ability to learn from incomplete data. Furthermore, some papers that have used these approaches are discussed below Table \ref{table.1}. 
\end{enumerate}
\newpage
\begin{table}[h!]
    \centering
    \caption{Various approaches of the gait recognition method used in different studies. The check mark indicates the approach used in the respective paper.}
    \label{table.1}
    \renewcommand{\arraystretch}{1.5}
    \resizebox{\textwidth}{!}{
        \begin{tabular}{|c|c|c|c|c|c|}
            \hline
            \Large{\textbf{Authors (et al.)}} & \multicolumn{2}{c|}{\Large{\textbf{Appearance-based approaches}}} & \multicolumn{3}{c|}{\Large{\textbf{Model-based approaches}}} \\
            \hline
            & \large{\textbf{Silhouette-Based}} & \large{\textbf{Non-Silhouette}} & \large{\textbf{2D Structural}} & \large{\textbf{3D Structural}} & \large{\textbf{Motion}} \\
            \hline
            \large{\textbf{Lwama \cite{iwama2012isir}}} & \large{\textbf{\checkmark}} & & & & \\
            \hline

            \large{\textbf{Bashir \cite{bashir2009gait}}} & \large{\textbf{\checkmark}}&&&&	\\
            \hline
            \large{\textbf{Hayfron-Acquah \cite{hayfron2003automatic}]}}&  \large{\textbf{\checkmark}}&&&&	\\
            \hline
            \large{\textbf{Sivapalan \cite{sivapalan2011gait}}}& \large{\textbf{\checkmark}}	&&&&	\\
            \hline
            \large{\textbf{Li \cite{li2017deepgait}}}& \large{\textbf{\checkmark}}	&&&&\\
            \hline
            \large{\textbf{Wu \cite{wu2016comprehensive}}}& \large{\textbf{\checkmark}}&&&&	\\
            \hline
            \large{\textbf{Zeng \cite{zeng2016view}}}&\large{\textbf{\checkmark}} &&&&\\
            \hline
            \large{\textbf{Kusakunniran \cite{kusakunniran2014recognizing}}} & &	\large{\textbf{\checkmark}}	&&&	\\
            \hline
            \large{\textbf{Brox \cite{brox2004high}}}&	& \large{\textbf{\checkmark}}	&&&	\\
            \hline
            \large{\textbf{Hu \cite{hu2012incremental}}}&	&\large{\textbf{\checkmark}} &&&\\
            \hline
            \large{\textbf{Akita \cite{akita1984image}}}&&&	\large{\textbf{\checkmark}}	&&	\\
            \hline
            \large{\textbf{Guo \cite{guo1994understanding}}}&&& \large{\textbf{\checkmark}} &	&		\\
            \hline
            \large{\textbf{Rohr \cite{rohr1994towards}}}&&& \large{\textbf{\checkmark}} &&	\\
            \hline
            \large{\textbf{Karaulova \cite{karaulova2000hierarchical}}} &&& \large{\textbf{\checkmark}} &&	\\
            \hline
            \large{\textbf{Niyogi \cite{niyogi1994analyzing}}}&&& \large{\textbf{\checkmark}} &&	\\
            \hline
            \large{\textbf{Cunado \cite{cunado2003automatic}}}&&&&& \large{\textbf{\checkmark}} \\
            \hline
            \large{\textbf{Yam \cite{yam2004automated}}}&&&&& \large{\textbf{\checkmark }}	\\
            \hline
            \large{\textbf{Wagg \cite{wagg2004automated}}}&&&&& \large{\textbf{\checkmark }}	\\
            \hline
            \large{\textbf{Bouchrika \cite{bouchrika2016towards},\cite{bouchrika2006markerless}}}&&&&& \large{\textbf{\checkmark}} \\
            \hline
            \large{\textbf{Ariyanto \cite{ariyanto2012marionette}}}&&&& \large{\textbf{\checkmark}} &	\\
            \hline
            \large{\textbf{Zhao \cite{zhao20063d}}}&&&& \large{\textbf{\checkmark}} &	\\
            \hline
            \large{\textbf{Tang \cite{tang2016robust}}}&&&& \large{\textbf{\checkmark}} &	\\
            \hline
            \large{\textbf{López-Fernández \cite{lopez2016new}}}&&&& \large{\textbf{\checkmark}} &		\\
            \hline
            \large{\textbf{Kastaniotis \cite{kastaniotis2016gait}}}&&&& \large{\textbf{\checkmark}} &		\\
        \end{tabular}
    }
\end{table}

There are many other methods that can improve the GR processes such as parametric eigenspace transformation, canonical space analysis, and deep learning method \cite{mogan2022advances}. 

\subsection{Parametric Eigenspace Transformation}

Parametric eigenspace transformation based on PCA \cite{shuai2021presentation}, is an effective method to accelerate GR and identification methods \cite{iwashita2021speed}. It can detect differences in gait patterns from an image sequence \cite{zhang2021cross}. This way, it is possible to recognize different gaits without using a database. It is also helpful for automatic attendance systems.  It is an effective metric for automatic gait analysis and faces recognition \cite{khalid2021npt}. The transformation reduces the dimensionality of data by transforming each image template from a high-dimensional to a low-dimensional space \cite{mc2021impact}. Gait is considered a sequence of images providing rich perceptual information \cite{chai2022multi}. Moreover, it can reveal a walker's identity, gender, and emotional state \cite{lee2022identification}. Therefore, it is essential to identify the gait signature consistent across multiple viewing conditions \cite{matasa2021deep}.

\subsection{Canonical Space Analysis}

Canonical space analysis is a valuable technique for reducing the dimensionality of data \cite{wu2021locality}. It is a powerful technique that can be used in GR methods. It works by transforming each image template from a high to a lower-dimensional canonical space \cite{tang2021dynamic}. This transformation makes GR easier and faster. This technique can achieve high recognition accuracy even for relatively small data sets \cite{amrani2021new}. It is also capable of handling changing walking conditions and camera viewing angles \cite{karthikram2021automatic}. The very demanding method for researchers nowadays is a deep neural network which is discussed below.

\section{Overview of Deep Learning}

DL is a subset of ML that uses many layers of neural networks. The first papers about DL were published in the 1980s \cite{lecun1989backpropagation}, but today it has become increasingly important in computer science, especially artificial intelligence \cite{janiesch2021machine}. Among the most notable achievements of this technology are computer programs that can play better than any human, speech recognition systems that are better than those used by humans, and cars that can drive themselves \cite{parloff2016deep}. DL is based on artificial neural networks \cite{fan2022copy,sharif2022decision}. DL algorithms are generally trained using massive amounts of data, i.e. image datasets, sound recordings, or data about human behavior \cite{bravo2021bioacoustic}. There is a large number of deep learning techniques used by researchers in present years \cite{arsalan2022prompt, iqbal2022g,aslam2022ensemble, khan2020machine, khan2022t, khan2021leveraging, umer2022comprehensive}.

\subsection{Recognizing Human Gait using Deep Learning and Computer Vision}
DL is used for various computer vision tasks, i.e. object detection and image captioning \cite{elguendouze2022towards}. These algorithms are trained to learn from data and perform specific tasks \cite{shao2021organic}. They can be used for image recognition, GR, and other visual recognition tasks \cite{ding2022interval}.\\

Computer Vision (CV) deals with enabling computers to see and understand visual content \cite{sarker2022ai}. It is a vast subject that can be applied to various fields, including medical science, food security, traffic monitoring, robotics, and human motion recognition \cite{danaci2022survey}. There are several ways that computer vision can be implemented. The most common are image recognition, object detection, and image segmentation. Image recognition is the ability to look at an image and understand it \cite{khan2021machine}. Object detection is the capability to identify individual objects in an image, even if the objects are incomplete or partially obstructed from view \cite{li2022analysis}. Segmentation allows you to  distinguish a segment into parts that you can describe individually.
 
In \cite{sahu2020contemporary}, authors presented a thorough understanding of the gait recognition framework to enhance the accuracy of several gait databases.
Palla et. al. \cite{palla2022human} proposed a novel FTS method that utilized the Firefly Algorithm to choose boundaries. PCA is used to reduce dimensionality, and multiple discriminant analysis (MDA) is used to increase class separability. The proposed technique was tested on a CASIA-B database, with the results showing excellent performance compared to other gait recognition systems discussed in published literature.
Rao et. al. \cite{rao2022adaptive} proposed an algorithm that utilized a Gait Energy Image as a template to select key features from the inputted gait data. Afterward, these selected traits are submitted to Principal Component Analysis (PCA) and Multi-Class Linear Discriminant Analysis (LDA) to enhance gait recognition accuracy. The proposed approach can markedly improve the efficacy of gait recognition compared to existing SOTA techniques.
  
\subsection{Application of Deep Learning to Aging or Disabled Subjects}
A DL model can be trained to recognize the faces of aging or disabled subjects \cite{thilagaraj2021electrooculogram}. One popular CNN model is the orthogonal Embedding-CNN model \cite{wang2018orthogonal}. This method uses two CNNs instead of one for face recognition. Some main steps are required to perform human GR tasks which is discussed below \cite{lee2002gait}: 

\subsubsection{Feature Selection}	

Currently, researchers have been working on the subject of human GR using DL \cite{prakash2018recent, sharif2020framework}. The idea is to create a system that can recognize human gaits from video. The process of recognizing human gaits is based on a deep neural network (DNN) and the features of a person's gait. The DNN can be trained to recognize human gait patterns and is highly effective in identifying human style \cite{zaki2020study}.

\subsubsection{Classification}
Almost all artificial neural network architectures trace back to similar input relevance values \cite{holzinger2021towards} means it can predict based on the same input features it picked up during training. This is an important aspect when studying human gait, as the symmetries of the left and right body movements can help identify an individual \cite{yan2021gait}. Furthermore, this approach can automatically classify disease and gait disorders \cite{nawaz2022hand,khan2022skin,khan2020lungs}.
\subsubsection{Accuracy}
Several factors must be considered to ameliorate the accuracy of human GR using DL \cite{addabbo2021temporal}. One of these factors is the amount of variability in the data. This variability is inherent to human movements. In addition, it is essential to build a robust and reliable model \cite{liu2005effect}.

\subsection{Limitations and challenges  of Deep Learning for Human Gait Recognition}

The following are some limitations of applying DL for GR \cite{alotaibi2017improved}. It may not work well,
\begin{enumerate}
    \item In low-light or dark environments. \\
    \item In noisy environments with much background noise. \\
     \item In environments where the person is wearing a heavy object or an outfit that affects how they walk. \\
     \item If the person is walking in a way that is not natural.\\ 
      \item If the person is walking unusually or oddly. \\
      \item  If the person is walking at a different speed from the sample that was used to train the model. \\
    \item If the person is walking in a different pose or posture than the sample for model training. \\
      \item If the person has undergone significant changes since the sample was taken, such as losing weight, gaining weight, etc.
      \end{enumerate}

Human gait recognition is a promising biometric for human identification, but it still faces challenges due to factors such as the intrinsic variability of human walking styles, environmental factors, clothing, and lighting conditions \cite{sethi2022comprehensive}. This technology's difficulty is its ability to accurately identify humans from movement patterns alone, which is also affected by age differences and disability status \cite{chen2022gait}. \\
Furthermore, algorithms are often required to be robust enough to handle multiple views of a person's walking style over time and across different environments \cite{parashar2022intra}. Finally, gait recognition algorithms must effectively deal with outliers - people who walk differently than most others – without negatively impacting accuracy \cite{trabassi2022machine}.

\section{Datasets}

DL (DL) is a powerful approach to ML to design the connection between inputs and outputs. The key advantage of DL is that it can be trained very efficiently using large datasets \cite{alzubaidi2021review}. This can be used for various applications, including recognizing people at airports or helping robots walk-in environments with rugged terrain \cite{chen2021semi}. 

\subsection{OU-ISIR LP Bag Dataset}

The OU-ISIR LP Bag dataset was first introduced in 2018 and has attracted considerable attention from the community \cite{uddin2018isir}. In less than two years, 18\% of GR methods have utilized it \cite{kececi2020implementation}. However, there are some caveats to using it. First of all, the dataset only includes gait data with carried objects. Second, it is intended for specific applications, such as single-viewpoint applications \cite{seely2008university}. The OU-ISIR dataset contains the gait silhouettes of 4,007 subjects. The subjects are aged from 0 to 94 years and span a wide age range. The data were acquired using two acquisition sessions with each subject. The subjects were recorded from 14 angles, with a 15-degree angle change at every step. The dataset has been extensively tested using cross-view testing protocols \cite{makihara2012isir}. It comprises two databases: the OU-ISIR Gait Database and the CASIA Gait Database B \cite{xu2017isir}. It has a population of over 4000 subjects, sufficient to test and improve the algorithms. The data are also used for age estimation.
\\

\begin{table}[h]
	
		\centering
		\caption{Human Gait publically available datasets with different approaches \\.}\label{table.2}
	\renewcommand{\arraystretch}{3}
		\resizebox{1\textwidth}{!}{
		
		\begin{tabular}{ |c|c|c|c|c| }
	
\hline		

\Large{\textbf{Authors (et. al.)}} &	\Large{\textbf{Year}}  &	\Large{\textbf{Approaches}} &	\Large{\textbf{Subjects}} & \Large{\textbf{Datasets}}\\
\hline	
\large{\textbf{Chalidabhongse \cite{chalidabhongse2001umd}}} & \large{\textbf{2001}} &	\large{\textbf{UMD}} &	\large{\textbf{55}} &	\large{\textbf{Video-Based  Dataset}}\\
\hline	
\large{\textbf{Phillips \cite{phillips2002baseline}}} & \large{\textbf{2002}} &	\large{\textbf{NIST}} &	\large{\textbf{74}} &	\large{\textbf{Video-Based  Dataset}}\\
\hline	
\large{\textbf{Wang \cite{wang2003silhouette}}} & \large{\textbf{2003}} &	\large{\textbf{CASIA}} &	\large{\textbf{124}}	& \large{\textbf{Video-Based Dataset}}\\
\hline	
\large{\textbf{Kale \cite{kale2004identification}}} & \large{\textbf{2004}} &	\large{\textbf{CMU}} &	\large{\textbf{25}} &	\large{\textbf{Video-Based  Dataset}}\\
\hline	
\large{\textbf{Sarkar \cite{sarkar2005humanid}}} & \large{\textbf{2005}} &	\large{\textbf{USF}}	& \large{\textbf{122}} &	\large{\textbf{Video-Based Dataset}}\\
\hline	
\large{\textbf{Nixon \cite{nixon2006automatic}}} & \large{\textbf{2006}}	& \large{\textbf{Small/large}} &	\large{\textbf{12}} &	\large{\textbf{Video-Based Dataset}}\\
\hline	
\large{\textbf{Hofmann \cite{hofmann2014tum}}} & \large{\textbf{2014}} &	\large{\textbf{TUM GAID}} &	\large{\textbf{305}}	& \large{\textbf{Video-Based Dataset}}\\
\hline	
\large{\textbf{Mantyjarvi \cite{mantyjarvi2005identifying} }}& \large{\textbf{2005}}	& \large{\textbf{Speed}}	 & \large{\textbf{36}} &  \large{\textbf{	Accelerometer-Based Dataset}}\\
\hline	
\large{\textbf{Gafurov \cite{gafurov2007gait}}} & \large{\textbf{2007}} &	\large{\textbf{Motion-recording}} &	\large{\textbf{50}} &	\large{\textbf{Accelerometer-Based Dataset}}\\
\hline	
\large{\textbf{Casale \cite{casale2012personalization}}} & \large{\textbf{2012}} &	\large{\textbf{Walking pattern}}	& \large{\textbf{22}} &	\large{\textbf{Accelerometer-Based Dataset}}\\
\hline	
\large{\textbf{Muaaz \cite{muaaz2012influence}}} & \large{\textbf{2012}} &	\large{\textbf{Android phone Google G1}} &	\large{\textbf{51}} &	\large{\textbf{Accelerometer-Based Dataset}}\\
\hline	
\large{\textbf{Ngo \cite{ngo2014largest}}} & \large{\textbf{2014}} &	\large{\textbf{Inertial sensor}} &	\large{\textbf{744}}	& \large{\textbf{Accelerometer-Based Dataset}}\\
\hline	
\large{\textbf{Reyes \cite{reyes2016transition}}} & \large{\textbf{2016}} &	\large{\textbf{Postural transitions}} &	\large{\textbf{30}}	& \large{\textbf{Accelerometer-Based Dataset}}\\
\hline	
\large{\textbf{Orr \cite{orr2000smart}}} & \large{\textbf{2000}} &	\large{\textbf{First floor-sensor}}	& \large{\textbf{15}} &	\large{\textbf{Floor-Sensor-Based Dataset}}\\
\hline	
\large{\textbf{Suutala \cite{suutala2004towards}}} & \large{\textbf{2004}} &	\large{\textbf{Footsteps of both feet}} &	\large{\textbf{11}}	& \large{\textbf{Floor-Sensor-Based Dataset}}\\
\hline	
\large{\textbf{Middleton \cite{middleton2005floor}}} & \large{\textbf{2005}} &	\large{\textbf{Walk without footwear}} &	\large{\textbf{15}}	& \large{\textbf{Floor-Sensor-Based Dataset}}\\
\hline	
\large{\textbf{Jenkins \cite{jenkins2007using}}} & \large{\textbf{2007}} &	\large{\textbf{Children}} &	\large{\textbf{62}} &	\large{\textbf{Floor-Sensor-Based Dataset}}\\
\hline	
\large{\textbf{Otero \cite{otero2005application}}} & \large{\textbf{2005}} &	\large{\textbf{First wave-radar}} &	\large{\textbf{49}}	&  \large{\textbf{Radar-Based Dataset}}\\
\hline	
\large{\textbf{Wang \cite{wang2011micro}}} & \large{\textbf{2011}} &	\large{\textbf{One-arm or two-arm}} &	\large{\textbf{1}} &	\large{\textbf{Radar-Based Dataset}}\\
\hline	
	\end{tabular}	
}
\end{table}
Some well-known publically available datasets to evaluate different approaches for Human GR systems are described in table \ref{table.2}. The number of subjects is also mentioned in the table above.

\section{Deep learning-based Approaches for Human Gait Recognition}

Every human gait is unique and challenging to identify by computer vision algorithms. One of the best ways to implement CNN is with Google’s pre-trained Tensorflow model called ‘Inception ResNet V2’ (or just ‘ResNet V2’) \cite{peng2022research}. There are various DL Models have been explored for GR. CNN and RNN (Recurrent Neural Networks) are ML models commonly used in computer vision tasks \cite{ward2021computer}. For Human GR, CNN and RNN are used because they can accurately recognize patterns in images. CNN detects patterns like shapes and their Orientations \cite{guindel2017joint}, whereas RNN can detect patterns like sequential order of shapes \cite{sharma2018csgnet}. CNN and RNN are used for Human GR in the following ways. CNN for GR with the image as input CNN for GR with video as input RNN for GR with sequential data as input. In DL, a model comprises several layers of nodes—neurons that recognize data patterns by comparing them to existing models. Deep networks have several advantages over traditional supervised ML methods \cite{hong2020machine}.\\\\
First, they can be trained to learn complex tasks more efficiently by leveraging their inherent representation strength. Second, they can effectively generalize across different domains by adapting their architecture and training procedure to the particular task. Third, they can be employed for large-scale assessment and forecasting on datasets too big or intricate for classic management strategies. The ability to recognize human gait could be advantageous in various applications. For instance, it could permit people with disabilities to traverse hindrances or regulate their wheelchairs. Moreover, it could enhance automated vehicles’ proficiency by enabling them to be more accurately detected as pedestrians. \\\\
Visual GR is used as one approach to train neural networks to recognize human gait. This entails taking pictures of people walking and using a deep-learning model to analyze the images. This approach is advantageous due to its relatively low cost and ease of implementation; however, there are some drawbacks \cite{jia2021cjam}\cite{liao2020model}. For instance, individuals must constantly film and upload video footage of their surroundings, which could be problematic in specific scenarios. Additionally, this method does not provide information about the person's health condition or other factors that might influence gait. Researchers have created alternative approaches to tackle these issues that combine image processing with extra features extracted from videos. These additional elements enable them to gain insight into an individual's health status or activity level to enhance the accuracy of the prediction model  \cite{muhammad2021human}.\\\\
Khan et. al. \cite{khan2020hand} proposed that the HAR system is a novel approach that combines conventional hand-crafted features with histograms of oriented gradients (HoG) and in-depth features. In the first phase, a human silhouette is detected using a saliency-based method. This model achieved excellent results as compared to the existing techniques. Das et. al. \cite{das2022unified} proposed an improved CNN architecture (WMsCNN). A weight update subnetwork (Ws) is designed to adjust the weights of certain features based on how much of a contribution they made to the final classification task. The weight-updated method is used by lowering these features' sensitivity to covariate factors. Global features are subsequently generated from these factors as a result of fusion.  Chiu et. al . \cite{chiu2022human} presented a method for GR using a point cloud Using Light Detection and Ranging (LiDAR) technology to detect human gaits. DL architecture is then used to improve human GR accuracy, which handles time-series data. Bayat et. al. \cite{bayat2022human} proposed method developed a bag-of-words feature representation based on a GR procedure. Its performance is  evaluated by comparing the classification results with extracted features, using human gait data collected from 93 individuals walking comfortably between two endpoints in two separate sessions. The proposed technique produced far more accurate classification results than standard statistical features in all the used classifiers.\\\\
Shopon et. al. \cite{shopon2022multi} introduced a new approach for recognizing people walking down an unconstrained pathway. The architecture input is joint body coordinates and an adjacency matrix representing the skeleton joints. The graph neural network framework incorporates a residual connection to smooth the input feature. This framework used kinematic relations, spatial and temporal variables, and joint body variables to identify gait. State-of-the-art GR methods outperformed the study's proposed method on unconstrained paths. The CASIA-B and multi-view gait AVA datasets assessed the method's efficacy. Li et. al. \cite{li2022gaitslice} proposed a new method, GaitSlice, to capture Spatio-temporal slice features from human motion. Slice Extraction Device (SED) develops slice features from the top down. For example, the residual frame attention mechanism (RFAM) focuses on keyframes. In order to mimic real life more closely, GaitSlice combines parallel RFAMs with slice information to identify critical Spatio-temporal components. The method is evaluated on CASIA-B and OU-MVLP datasets and compared with six GR algorithms, taking rank-1 accuracy as the benchmark. Also compared, were the results with cross-view and walking circumstances. In this paper \cite{10.1007/s11063-021-10709-1}, the authors proposed a residual structure to preserve more identity information while implementing the view transformation. A fusion model is used to combine the outcomes of the three perspectives at the recognition and decision-making stage to assess the method's effectiveness in this paper. The CASIA-B gait data set was employed to assess the algorithm's performance in this study. This model performed better than prior networks and can detect more accurately than prior models, primarily when clothing and viewing angles differ.\\\\

\begin{sidewaystable}

		\centering
		\caption{DL-based approaches and datasets for human gait recognition.\\}\label{table.3}
	\renewcommand{\arraystretch}{4}
		\resizebox{1\textwidth}{!}{
		
		\begin{tabular}{ |c|c|c|c|c|c| }
	
\hline		
\Large{\textbf{Ref No.}} &	\Large{\textbf{Year}}  & \Large{\textbf{Approaches}} &	\Large{\textbf{Methods}} & \Large{\textbf{Datasets}} & \Large{\textbf{Results(Acc.\%)}}\\

\hline	
\Large{\textbf{\cite{arshad2022multilevel}}} & \Large{\textbf{2022}} &	\Large{\textbf{Multilevel paradigm for CNN features selection}} & \Large{\textbf{Classification}} &	\Large{\textbf{AVAMVG gait, CASIA A, B and C}} &\Large{\textbf{V99.8, 99.7, 93.3 and 92.2}}\\
\hline

\Large{\textbf{\cite{arshad2019multi}}} & \Large{\textbf{2019}} &	\Large{\textbf{Multi-level features fusion and selection}} &	\Large{\textbf{Selection}} &	\Large{\textbf{AVA multi-view gait (AVAMVG), CASIA A, B and C}} & \Large{\textbf{100, 98.8, 87.7, and 91.6}}\\
\hline	

\Large{\textbf{\cite{mehmood2020prosperous}}} & \Large{\textbf{2020}} &	\Large{\textbf{Feature selection by Firefly algorithm and Skewness}} &	\Large{\textbf{Selection}}	& \Large{\textbf{180, 360 and 540 degrees angles of CASIA B}} & \Large{\textbf{94.3, 93.8 and 94.7}}\\
\hline
	
\Large{\textbf{\cite{khan2021human}}} & \Large{\textbf{2021}} &	\Large{\textbf{Euclidean Norm and Geometric Mean Maximization}} &	\Large{\textbf{Selection}}	& \Large{\textbf{CASIA B}} & \Large{\textbf{96.0}}\\
\hline
	
\Large{\textbf{\cite{saleem2021human}}} & \Large{\textbf{2021}} & \Large{\textbf{Mean absolute deviation extended serial fusion}} &	\Large{\textbf{Classification}}	& \Large{\textbf{CASIA B}} & \Large{\textbf{89.0}}\\

\hline
			
\Large{\textbf{\cite{khan2022human}}} & \Large{\textbf{2022}} &	\Large{\textbf{Improved Ant Colony Optimization}} &	\Large{\textbf{Classification}}	& \Large{\textbf{0, 18, and 180 degrees angles of CASIA B}} & \Large{\textbf{95.2, 93.9, and 98.2}}\\
\hline
			
\Large{\textbf{\cite{sharif2022deep}}} & \Large{\textbf{2022}} &	\Large{\textbf{Kurtosis-controlled entropy, Resnet 101}} &	\Large{\textbf{Selection and Classification}}	& \Large{\textbf{CASIA B and Real time captured videos}} & \Large{\textbf{95.2 and 96.6}}\\
\hline

\Large{\textbf{\cite{mehmood2022human}}} & \Large{\textbf{2022}} &	\Large{\textbf{	VGG-16, PSbK and OAMSVM}} &	\Large{\textbf{Selection and Classification}}	& \Large{\textbf{	00◦, 18◦, 36◦, 54◦, 72◦, and 90◦ of CASIA B}} & \Large{\textbf{95.8, 96.0, 95.9, 96.2 and 95.6}}\\
\hline
				
\Large{\textbf{\cite{khan2022human}}} & \Large{\textbf{2022}} &	\Large{\textbf{Improved moth-flame optimization}} &	\Large{\textbf{Classification}}	& \Large{\textbf{CASIA B and TUM GAID}} & \Large{\textbf{	91.20 and 98.60}}\\
\hline
					
\Large{\textbf{\cite{sharif2020machine}}} & \Large{\textbf{2020}} &	\Large{\textbf{Threshold Based Feature Fusion and Multi-class SVM}} &	\Large{\textbf{Selection}}	& \Large{\textbf{CASIA A, B and C}} & \Large{\textbf{98.6 ,93.5 and 97.3}}\\
\hline

\Large{\textbf{\cite{nasir2021human}}} & \Large{\textbf{2021}} &	\Large{\textbf{ML and Neuro Fuzzy Classifier}} &	\Large{\textbf{Classification}}	& \Large{\textbf{HMDB-51 and Hollywood2}} & \Large{\textbf{82.5 and 91.9}}\\
\hline

\Large{\textbf{\cite{nasir2022harednet}}} & \Large{\textbf{2022}} &	\Large{\textbf{Hybrid recognition technique}} &	\Large{\textbf{Selection}}	& \Large{\textbf{NTU RGB+D, HMDB51, and UCF-101}} & \Large{\textbf{97.4, 80.5, and 97.4}}\\
\hline

\Large{\textbf{\cite{deng2022human}}} & \Large{\textbf{2022}} &	\Large{\textbf{Seven layer fully connected layers}} &	\Large{\textbf{Fusion}}	& \Large{\textbf{CASIA B}} & \Large{\textbf{90.0}}\\
\hline

	\end{tabular}	
}

\end{sidewaystable}

\newpage

Table \ref{table.3} shows some DL-based approaches along with well-known datasets to evaluate the Human GR systems. The results of each method are compared in terms of accuracy.

\section{Conclusion}
GR (Gait Recognition) is an exciting application of neural networks and deep learning that has been seen in science fiction for a long time but is now starting to be applied in the real world. This is an inspiring example of how deep learning can be used for more than just image recognition. GR provides a challenge due to considering a range of variables like a person’s mood, age, fitness level, clothing type, and way of walking. Although not perfect, it still serves as a valuable tool with potential for further applications as technology improves. Physically or behaviorally measurable aspects of an individual called biometrics are utilized in this visual pattern recognition task which may be one of the most complex tasks in computer science. This makes it much easier to collect data for ML algorithms. GR is even more helpful because it is a passive biometric. This means it is automatically collected by visual sensors rather than requiring a person to act. There is a lot of research on GR in recent years and it is used in biometric technology. Although it has a high accuracy rate, it is hard to measure and validate the algorithm's performance. Some efforts have been made to push the boundaries of GR. This includes the use of the DL model, which can cognize the people based on more than one modality. This includes visual and acoustic data alongside the person's walking pattern. Various neural network architectures have been tested, including Long Short Term Memory (LSTM) networks, Convolutional Neural Networks (CNN), and GRU networks.
\section{Future Direction}
Deep Learning (DL) is a widely-used Artificial Intelligence (AI) technique for Human Gait Recognition (GR). DL algorithms, such as deep neural networks, are trained to detect patterns from large datasets. Recent breakthroughs in DL have enabled the recognition of complex concepts, such as human gait. However, there are certain areas for improvement with existing DL methods that restrict their application to Human GR. One limitation is the lack of ground truth; it can be difficult to determine whether an object is a person. Another challenge is low accuracy; current DL techniques often suffer from low accuracy levels, particularly when trying to recognize humans. This makes them unsuitable for practical implementation in the real world. New ways of training and assessing DL models need to be developed to address these issues and make DL more suitable for real-world use.\\\\

Researchers are exploring using deep learning (DL) models to detect abnormalities in gait patterns. These systems can be taught to recognize typical ailments such as Parkinson's or stroke by analyzing video footage of people walking normally. By distinguishing between normal and abnormal motion patterns, it may be possible to detect cases that require urgent medical attention. With DL, computers can process and interpret human movement from videos and images; thus enabling them to comprehend people's gestures and activities, making them more interactive and responsive. A primary challenge in this field is training abundant data for deep learning models, which could originate from different sources like motion capture systems, video recordings, or treadmill tests. Consequently, obtaining high-grade data from various sources is essential for guaranteeing robustness and precision when applying deep learning methods to human gait recognition. Another difficulty is using deep learning architectures for large-scale human gait identification. 
\section*{Conflict of Interest Statement}
\label{sec:conflict-of-interest}
The authors declare that they have no conflicts of interest related to this work. The research was conducted in the absence of any commercial or financial relationships that could be construed as a potential conflict of interest.

\bibliography{sn-bibliography}


\end{document}